\pdfoutput=1

\documentclass[11pt]{article}

\usepackage[preprint]{acl}

\usepackage{times}
\usepackage{latexsym}

\usepackage[T1]{fontenc}

\usepackage[utf8]{inputenc}

\usepackage{microtype}

\usepackage{inconsolata}

\usepackage{graphicx}
\usepackage{subfigure}
\usepackage{multirow}
\usepackage{multicol}
\usepackage{booktabs}
\usepackage{amssymb}
\usepackage{bbding}
\usepackage{amsmath}
\usepackage{amsfonts}
\usepackage{amssymb}
\usepackage{hyperref}
\usepackage{threeparttable}
\usepackage{pifont}
\usepackage{listings}

\title{BLSP-KD: Bootstrapping Language-Speech Pre-training \\ 
via Knowledge Distillation}

\author{Chen Wang\textsuperscript{1,3}\thanks{Work was done while at Alibaba Group.}, Minpeng Liao\textsuperscript{2}, Zhongqiang Huang\textsuperscript{2}\thanks{Corresponding author.}, Jiajun Zhang\textsuperscript{1,3}\footnotemark[2] \\
\textsuperscript{1} Institute of Automation, Chinese Academy of Sciences\\
\textsuperscript{2} Machine Intelligence Technology Lab, Alibaba Group.\\
\textsuperscript{3} School of Artificial Intelligence, University of Chinese Academy of Sciences \\
\texttt{\{wangchen2020\}@ia.ac.cn} \texttt{\{jjzhang\}@nlpr.ia.ac.cn} \\
\texttt{\{minpeng.lmp,z.huang\}@alibaba-inc.com} \\
}

\begin{document}
\maketitle
\begin{abstract}
Recent end-to-end approaches have shown promise in extending large language models (LLMs) to speech inputs, but face limitations in directly assessing and optimizing alignment quality and fail to achieve fine-grained alignment due to speech-text length mismatch. We introduce BLSP-KD, a novel approach for Bootstrapping Language-Speech Pretraining via Knowledge Distillation, which addresses these limitations through two key techniques. First, it optimizes speech-text alignment by minimizing the divergence between the LLM's next-token prediction distributions for speech and text inputs using knowledge distillation. Second, it employs a continuous-integrate-and-fire strategy to segment speech into tokens that correspond one-to-one with text tokens, enabling fine-grained alignment. 
We also introduce Partial LoRA (PLoRA), a new adaptation method supporting LLM finetuning for speech inputs under knowledge distillation. Quantitative evaluation shows that BLSP-KD outperforms previous end-to-end baselines and cascaded systems with comparable scale of parameters, facilitating general instruction-following capabilities for LLMs with speech inputs. This approach provides new possibilities for extending LLMs to spoken language interactions.

\end{abstract}

\section{Introduction}
The advent of Large Language Models (LLMs), such as ChatGPT~\cite{openai2023gpt}, Gemini~\cite{geminiteam2024gemini}, and various open-source models~\cite{bai2023qwen, touvron2023llama}, has revolutionized the field of natural language processing, showcasing unparalleled proficiency in language understanding and generation. These models, trained on vast text corpora, have demonstrated remarkable success across a broad spectrum of language tasks, ranging from traditional NLP tasks like summarization and translation, to conversational AI, code generation and interpretation, math problems solving, and to autonomous agents~\cite{zhang2023unifying, liu2023mathematical, wang2024survey}. As the capabilities of LLMs continue to evolve, there has been an increasing interest in extending their capabilities beyond text to include speech, a fundamental and intuitive form of human communication.

Speech, unlike text, conveys a wealth of information not only through its lexical content but also through paralinguistic features such as tone, emotion, and intent~\cite{busso2008iemocap, castro-etal-2019-towards}. Integrating the nuances of speech into LLMs presents significant challenges due to the modality difference. Cascaded systems, like those used in voice chat in ChatGPT and Gemini, where speech inputs are first converted to text using Automatic Speech Recognition (ASR), have been useful. However, these systems suffer from error propagation from the ASR component to the LLM, and cannot capture the rich paralinguistic information in speech due to the conversion to text.

Recent advancements have explored end-to-end approaches for aligning speech and text for LLMs, eliminating the need for separate ASR systems. These approaches, such as SpeechGPT~\cite{zhang2023speechgpt} and LLaSM~\cite{shu2023llasm}, leverage speech instruction data and aim to train models to enable instruction-following capabilities of LLMs to speech inputs. In particular, the recently proposed BLSP approach~\cite{wang2023blsp} introduced the concept of behavior alignment by instructing an LLM to generate continuations of speech transcripts and then using these continuations as supervised targets for training a speech adapter. While these models represent a progression towards unlocking the potential of end-to-end modeling, they face significant limitations. 

Firstly, these methods lack a direct measure for assessing and optimizing speech-text alignment quality. Secondly, the speech-text length mismatch hinders fine-grained alignment at the lexical level, which is essential for achieving accurate instruction-following capabilities. As a result, these approaches rely heavily on the availability and quality of speech instruction data, whether manually collected or automatically synthesized, and aim to maximize the likelihood of speech instruction data through a cross-entropy loss. However, such an approach could override the behavior of the LLM, which is carefully tuned through extensive supervised fine-tuning and reinforcement learning with human feedback, and cause a detrimental effect on the end-to-end models' alignment with human preferences, even if the LLM is kept frozen during the speech-text alignment process.

To address these challenges, we propose BLSP-KD, Bootstrapping Language-Speech Pretraining via Knowledge Distillation, a novel approach that employs two key techniques. First, we treat speech-text alignment as a knowledge distillation problem and use the KL-divergence of next-token prediction distributions between speech and text inputs as a direct measure of alignment. This measure can be optimized to ensure that the LLM's responses to speech inputs closely mirror those to corresponding text inputs. Second, we utilize a continuous-integrate-and-fire mechanism, integrated with transformer blocks, as the modality adapter to segment and project speech representations into hidden states with one-to-one correspondence with text tokens in the transcript. This enables measuring speech-text alignment at the token level, allowing for direct optimization of alignment using speech and transcripts alone, making it possible to perform speech-text alignment without speech instruction data. Additionally, we propose Partial LoRA (PLoRA), a new adaptation method that supports LLM finetuning for speech inputs under the knowledge distillation framework.

Through quantitative evaluations, we demonstrate that BLSP-KD outperforms both comparable cascaded systems and previous end-to-end baseline, BLSP, in terms of instruction-following capabilities for LLMs with speech inputs. While the focus of this paper is on understanding the lexical content in speech, we believe that the end-to-end modeling approach holds promise for the next step toward comprehending paralinguistic cues in speech, to fully leverage speech as a medium for interaction with LLMs.


\section{Background}
\label{sec:background}

We begin by formulating the problem definition. Consider an existing LLM trained solely on text, represented by parameters $\phi$. When prompted by a text input $\mathbf{x}$, this LLM can generate a response according to distribution $p_{\phi}(\cdot | \mathbf{x})$. Our objective is to extend the LLM's understanding and generation capabilities to speech inputs. Assume $\mathbf{x}$ is the transcript of a speech input $\mathbf{s}$. Typically, the speech is first transformed into feature representations $\mathbf{s}^\text{enc}$ by a speech encoder with parameters $\psi$, and then mapped through a modality adapter module with parameters $\theta$ to hidden states $\mathbf{s}^\text{adp}$ in the word embedding space, prompting the LLM to generate responses according to distribution $p_{\psi, \theta, \phi}(\cdot|\mathbf{s}^\text{adp})$. 




Previous methods address this problem by fine-tuning on a collection $\mathcal{D}=\{(\mathbf{s},\mathbf{y})\}$ comprised of pairs of speech input $\mathbf{s}$ and the desired text response $\mathbf{y}$. Given the limited availability of manually annotated speech instruction datasets, the recent BLSP approach suggests to align speech and text via behavior alignment. It constructs synthesized responses from an existing ASR dataset $\mathcal{D} = \{(\mathbf{s},\mathbf{x})\}$ of speech and transcript pairs using a continuation prompt $\mathbf{c}$, expanding it to a collection of tuples $\mathcal{D} = \{(\mathbf{s},\mathbf{x}, \mathbf{y})\}$ with each $\mathbf{y}$ generated via greedy-search as follows:
\begin{align}
\mathbf{y} = \arg\max_{\mathbf{y}'} p(\mathbf{y'} |\mathbf{x})
\end{align}

\noindent Here, and when clear from the context, the continuation prompt $\mathbf{c}$ is used as a condition but is not explicitly included to maintain clarity in presentation. 

The parameter set $\theta$ of the modality adapter module is then optimized to minimize the cross entropy loss of responses as in instruction fine-tuning:

\begin{align}
\ell^\text{resp}_\text{CE} (\mathbf{s}, \mathbf{y}) &=-\log p_\theta(\mathbf{y}|\mathbf{s}^\text{adp}) \nonumber \\ 
&= -\sum_j\log p_\theta(y_j|\mathbf{s}^\text{adp}, \mathbf{y}_{<j}) \label{eq:resp_ce}
\end{align}

\begin{figure*}[ht]
\centering
\includegraphics[width=0.9\textwidth]{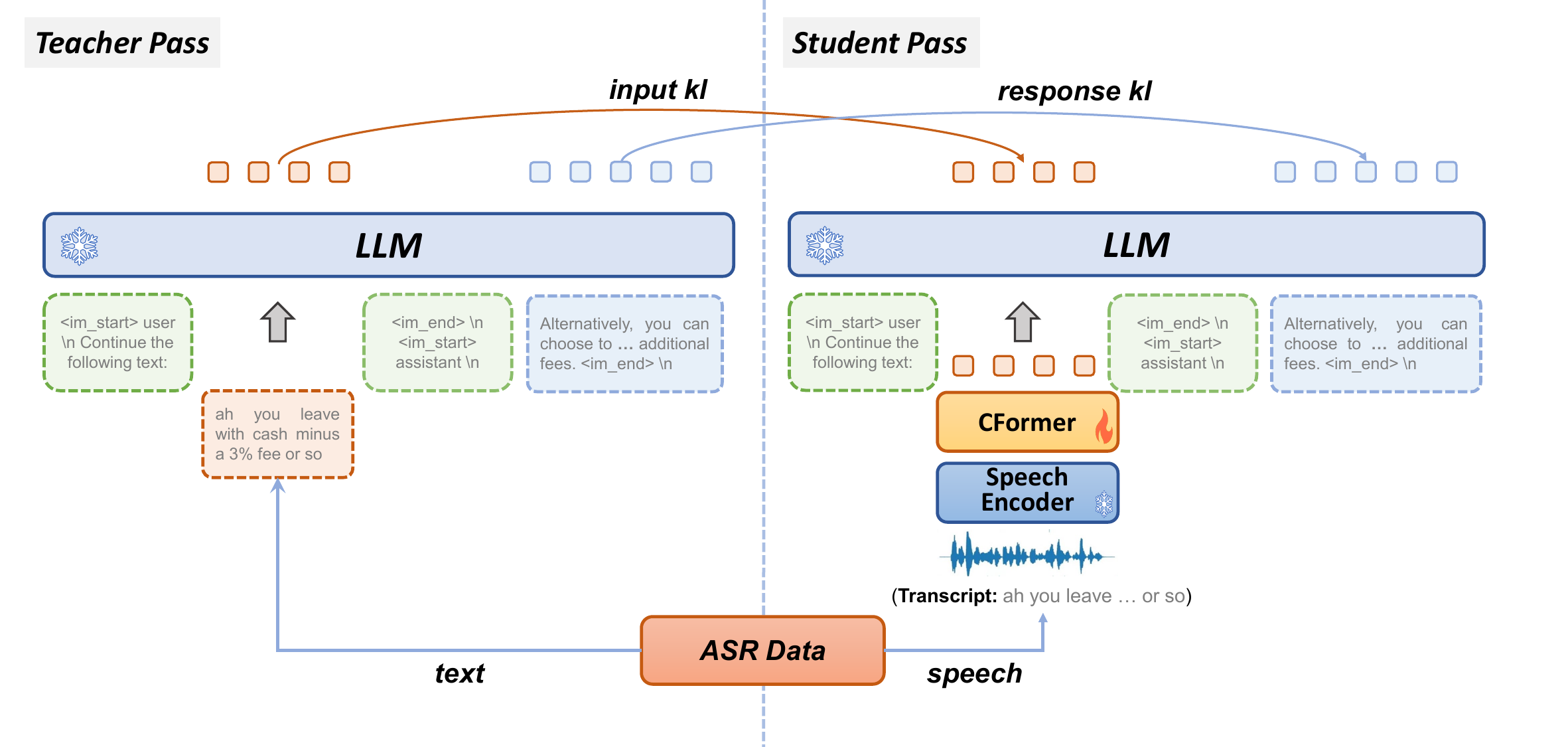}
\caption{An overview of the BLSP-KD approach. The next token prediction distributions based on text input, as computed by the LLM in the teacher pass on speech transcripts, are used as supervision in the student pass to train the modality adapter (CFormer). This allows the LLM to produce similar next token prediction distributions for both the input (orange) and response (blue) tokens given speech input. Note that in this figure, the LLM and Speech Encoder are kept frozen, as this is the setting used in most experiments; however, they can also be fine-tuned.}
\label{fig:blsp-kl}
\end{figure*}

Despite achieving remarkable instruction-following capabilities for speech inputs, the BLSP \citep{wang2023blsp} approach encounters two limitations. First, while the continuation instruction aims to generate diverse responses that mimic general next-token prediction, it represents only one specific instruction and thus captures only a subset of the LLM's generative behavior. Second, the cross-entropy loss encourages the model to replicate the 1-best responses with high probability and thus neglect the broader distribution of possible responses. The cross-entropy loss $\ell^\text{resp}_\text{CE} (\mathbf{s}, \mathbf{y})$, defined between speech $\mathbf{s}$ and 1-best response $\mathbf{y}$, does not provide an end-to-end differential measure of alignment quality between speech $\mathbf{s}$ and transcript $\mathbf{x}$.

\section{Method}
\label{sec:method}

\subsection{Knowledge Distillation}
In our proposed BLSP-KD approach, we consider speech-text alignment from the perspective of knowledge distillation. As shown in Figure~\ref{fig:blsp-kl}, given a tuple $(\mathbf{s}, \mathbf{x}, \mathbf{y})$, we treat the LLM's prediction distribution, $p(\cdot|\mathbf{x}, \mathbf{y}_{<j})$, of the next response token, after having observed the text input $\mathbf{x}$ and generated partial response $\mathbf{y}_{<j}=y_1, \cdots, y_{j-1}$, as the teacher distribution. In contrast, we consider the corresponding distribution, $p_\theta(\cdot|\mathbf{s}^\text{adp}, \mathbf{y}_{<j})$, for the speech input as the student distribution. If speech and text are well-aligned, the two distributions should be close to each other, as measured by KL divergence. This gives the following response KL loss: 

\begin{align}
&\ell^\text{resp}_\text{KL}(\mathbf{s}, \mathbf{x},\mathbf{y}) = \nonumber \\
&\quad -\sum_{j,y} p(y|\mathbf{x}, \mathbf{y}_{< j})\log p_\theta(y|\mathbf{s}^\text{adp}, \mathbf{y}_{< j}) \label{eq:resp_kl}
\end{align} 

This KL loss introduces a quantitative measure of speech and text alignment at each step of the response generation process. By minimizing this loss, we can learn a modality adapter for speech inputs that facilitates generation behaviors similar to those of text inputs when generating responses. 


\subsection{CFormer Adapter}

The modality adapters commonly found in literature compress the speech encoder's feature representations $\mathbf{s}^\text{enc}=s^\text{enc}_1,\cdots,s^\text{enc}_l$ to hidden states $\mathbf{s}^\text{adp}=s^\text{adp}_1,\cdots,s^\text{adp}_m$ at a fixed ratio, due to their use of convolutional or stacking methods~\citep{wang2023blsp, hu2024wavllm}. This typically results in the length of speech features $m$ not being equal to the corresponding number of text tokens $n$.


The discrepancy in length presents two challenges. First, text LLMs are pretrained at the granularity of text tokens, and the inconsistency between the granularity of speech and text inputs can complicate modality alignment. Second, the lack of one-to-one correspondence prevents measuring alignment between speech and text inputs at the prompt input stage of LLMs. Consequently, previous work has focused solely on alignment at the generation stage.

\begin{figure}[t]
\centering
\includegraphics[width=0.45\textwidth]{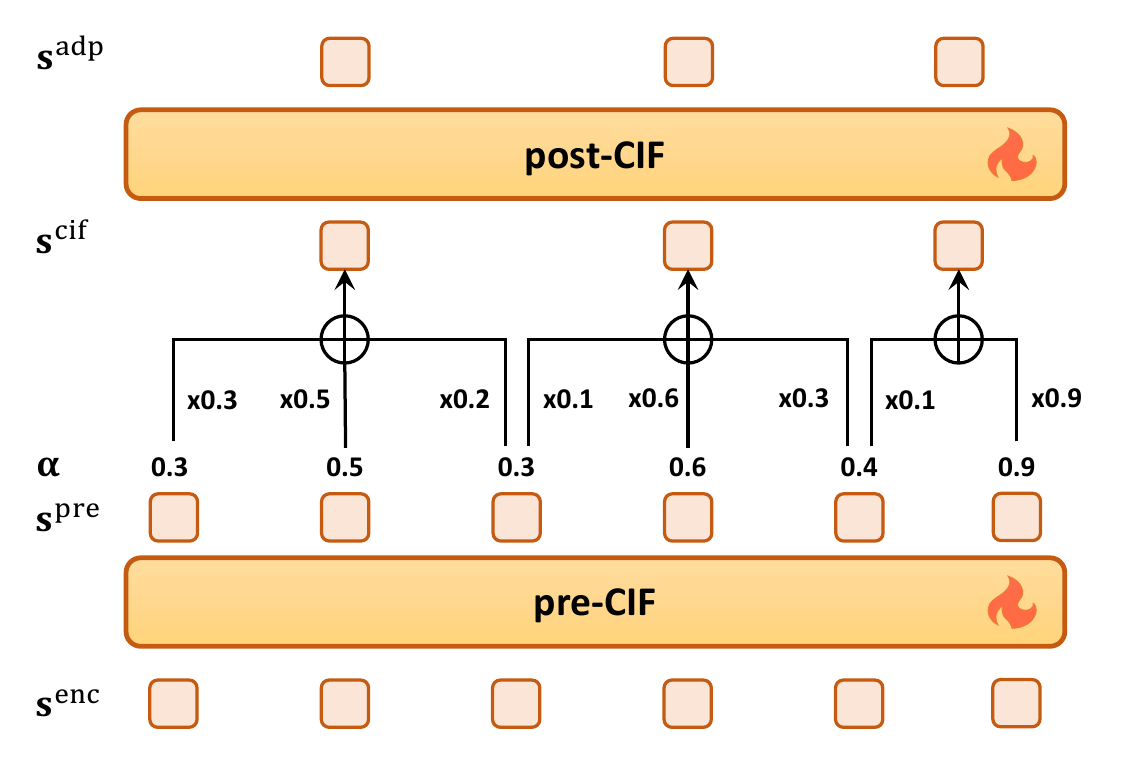}
\caption{An illustration of the CFormer adapter for mapping the speech encoder's feature representation $\mathbf{s}^\text{enc}$ of length $l$ to hidden states $\mathbf{s}^\text{adp}$ of length $n$ as inputs to the LLM. The $\alpha$ values are first computed for each of the $l$ hidden states in $\mathbf{s}^\text{pre}$ and then distributed among the number of text tokens to compute the $n$ hidden states in $\mathbf{s}^\text{cif}$.}
\label{fig:cformer}
\end{figure}

To address this issue, we employ the Continuous Integrate-and-Fire (CIF) mechanism \cite{dong2020cif} to dynamically segment speech representations into $n$ segments, each corresponding to a text token in the transcript. As a result, the speech hidden states $\mathbf{s}^\text{adp} = s^\text{adp}_1, \cdots, s^\text{adp}_n$ and the text tokens $\mathbf{x} = x_1, \cdots, x_n$ have a one-to-one correspondence at the token level. This allows us to apply knowledge distillation on the input token sequence using the input KL loss, as illustrated in Figure~\ref{fig:blsp-kl}:
\begin{align}
\ell^\text{input}_\text{KL}(\mathbf{s}, \mathbf{x}) = -\sum_{i,x} p(x|\mathbf{x}_{<i})\log p_\theta(x|\mathbf{s}^\text{adp}_{< i}) \label{eq:input_kl}
\end{align}

Unlike previous applications of the CIF mechanism in multimodal LLMs \cite{chen2023x} that primarily aimed at recognition accuracy, our approach focuses on measuring and optimizing speech-text alignment at the token level after being processed by LLMs.

Specifically, we introduce a CFormer adapter that incorporates a CIF block between two transformer blocks: a pre-CIF block processes the speech encoder's output $\mathbf{s}^\text{enc}$ to $\mathbf{s}^\text{pre}$, and a post-CIF block projects the CIF-generated hidden states $\mathbf{s}^\text{cif}$ to $\mathbf{s}^\text{adp}$ for input into the LLM, as illustrated in Figure~\ref{fig:cformer}.
The CIF block segments and projects hidden states $\mathbf{s}^\text{pre}=s^\text{pre}_1,\cdots,s^\text{pre}_l$ to $\mathbf{s}^\text{cif}=s^\text{cif}_1,\cdots,s^\text{cif}_n$, based on the CIF mechanism. Let $d$ be the feature dimension of the hidden states. The CIF block first determines the amount of information an input hidden state $s^\text{pre}_i$ represents with respect to a token from its last element $s^\text{pre}_{i,d}$ using a sigmoid function $\sigma$, optimized so that the sum matches the total number of tokens $n=|\mathbf{x}|$ in the transcript. The CIF loss is defined as follows:
\begin{align}
\ell_\text{CIF}(\mathbf{s}, \mathbf{x}) = \frac{|\sum_{i} \sigma(s^\text{pre}_{i,d}) - n|}{n} \label{eq:cif_loss}
\end{align}

During inference, $\alpha_i=\sigma(s^\text{pre}_{i,d})$ is used directly to segment the speech into tokens. At training time, however, the $\alpha$ values are first normalized to ensure that their sum equals the total number of tokens in the transcript:
\begin{align}
    \alpha_i &= \frac{\sigma(s^\text{pre}_{i,d})}{\sum_{i'}\sigma(s^\text{pre}_{i',d})} n  \nonumber
\end{align}

\noindent The $\alpha$ values are then accumulated from left to right to monotonically align the hidden state $s^\text{pre}_i$ to the $j$-th token with weight $\alpha_{i,j}$, ensuring that each $\alpha_i$ is distributed among one or multiple consecutive tokens and each token receives an accumulative weight of 1 from one or more consecutive input hidden states:  
\begin{equation}
    \sum_j \alpha_{i,j} = \alpha_i, \quad \sum_i \alpha_{i,j} = 1, \quad \alpha_{i,j} \geq 0 \nonumber
\end{equation}

The CIF block's output hidden states $\mathbf{s}^\text{cif}=s^\text{cif}_1,\cdots,s^\text{cif}_n$ are then computed as the weighted sum of the input hidden states $\mathbf{s}^\text{pre}=s^\text{pre}_1,\cdots,s^\text{pre}_l$:
\begin{align}
     s^\text{cif}_j & = M \sum_i \alpha_{i,j}s^\text{pre}_{i,1:d-1} \nonumber
\end{align}

\noindent where $s^\text{pre}_{i,1:d-1}$ excludes the last element in $s^\text{pre}_i$ and $M$ is a $(d-1)\times d$ projection matrix to maintain a dimension of $d$ for the hidden states.
Finally, the post-CIF block transforms $\mathbf{s}^\text{cif}$ into the final hidden states $\mathbf{s}^\text{adp}=s^\text{adp}_1,\cdots,s^\text{adp}_n$. 




\subsection{Partial LoRA}
\label{subsec:plora}

\begin{figure}[ht]
\centering
\includegraphics[width=0.45\textwidth]{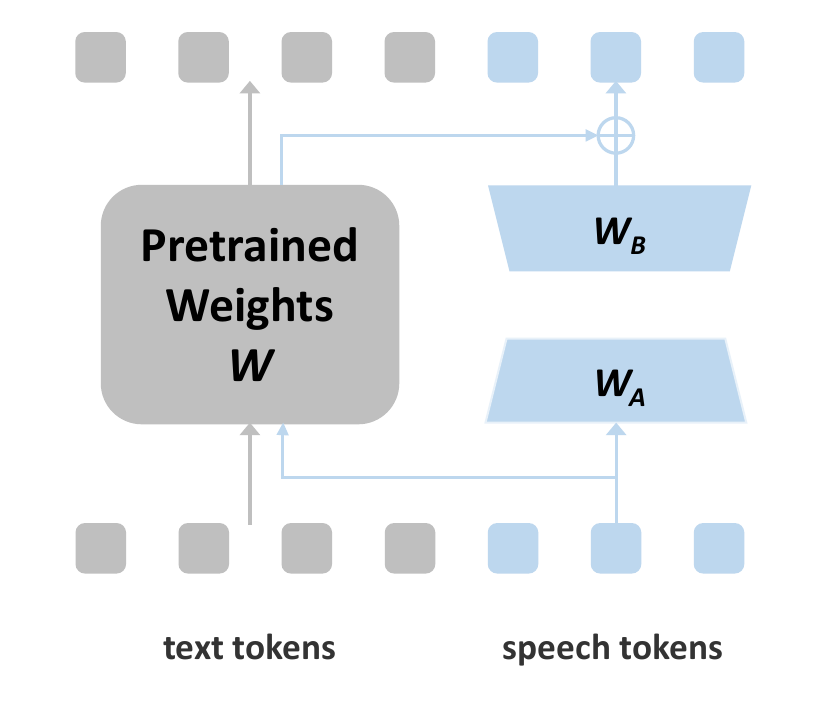}
\caption{The illustration of PLoRA. LoRA is only applied to speech tokens, while the encoding of text tokens remains unaffected.}  
\label{fig:plora}
\end{figure}

Speech inherently carries more information (e.g., paralinguistic cues), variations (e.g., different speakers and accents), and noise (e.g., background and channel noise). LLMs, trained solely on text tokens, lack the inherent ability to handle these discrepancies. The modality adapter, with its limited modeling power compared to LLMs, may be insufficient to bridge this gap fully. One solution is to unfreeze the LLM parameters to leverage its full modeling power for handling the complexities and variations of speech. Studies in multimodal models have demonstrated that unfreezing LLM parameters leads to more effective cross-modal alignment \cite{yin2023survey, zhang2024mm}.

However, applying fine-tuning methods like LoRA \cite{hu2021lora} is nontrivial within the knowledge distillation framework, as our goal is to maintain the LLM's behavior for text inputs while aligning its behavior for speech inputs to that of text inputs. Direct application of LoRA would change the LLM's behavior regardless of input modality, resulting in a degenerate solution that minimizes the KL loss but alters the LLM's behavior on text inputs as well.

To address this, we introduce Partial LoRA (PLoRA), designed to efficiently adapt LLM parameters only on partial inputs, i.e., the speech modality in our case. As depicted in Figure \ref{fig:plora}, PLoRA incorporates a low-rank adaptation only activated on the speech modality of the input tokens, providing the LLM additional modeling capacity to address the discrepancies between speech and text modalities, while preserving its  integrity in encoding and generating text tokens. A similar approach is also employed in \cite{dong2024interlm} to achieve efficient cross-modal alignment between image and text.

\section{Experiments}
\subsection{Model Configurations}
\label{sec:model_config}
For our proposed BLSP-KD approach, we employ Qwen-7B-chat \citep{bai2023qwen} as the large language model (LLM) and the encoder part of whisper-large-v2 \citep{radford2022robust} as the speech encoder, thanks to their demonstrated competitive performance among open source models. The CFormer adapter is composed of a 4-layer pre-CIF transformer block, a CIF block, and a 4-layer post-CIF transformer block, with the transformer blocks sharing the same architectural design and configuration as the whisper-large-v2 encoder.

In the majority of our experiments, we focus exclusively on tuning the parameters of the CFormer adapter, while keeping the speech encoder and the LLM frozen. Training is conducted over 3 epochs with a batch size of 96.
Additionally, we explored unfreezing all parameters of the speech encoder and unfreezing the LLM using PLoRA adaptation.

\subsection{Training Data}
\label{sec:training_data}
We utilize publicly available speech recognition datasets, including LibriSpeech \citep{panayotov2015librispeech}, GigaSpeech M set \citep{chen2021gigaspeech}, and Common Voice 13.0 \citep{commonvoice:2020}, resulting in approximately 1.9 million (speech, transcript) pairs encompassing 3k hours audio recordings.

As outlined in Section~\ref{sec:method}, our proposed BLSP-KD approach enables training directly on ASR training data, utilizing speech-transcription pairs with the input KL loss described in Eq~\ref{eq:input_kl}, in addition to training on speech instruction data using the response CE or KL losses described in Eq~\ref{eq:resp_ce} and Eq~\ref{eq:resp_kl}, respectively. Given the challenge of collecting speech instruction data in large volumes, we adopt the strategy detailed in BLSP~\cite{wang2023blsp} to generate synthesized speech instruction data through a continuation writing prompt. Note that computing the response CE or KL losses necessitates full context, including the continuation prompt, whereas the input KL loss does not require any prompt. However, when used in conjunction with the response CE or KL losses, we incorporate the continuation prompt as a condition for computing the input KL loss, allowing for a single forward pass on a minibatch.

\subsection{Baselines}

We compare our method with the following baselines.

\paragraph{Text+LLM} The input to the LLM is the ground-truth speech transcripts. 

\paragraph{Whisper+LLM} The input to the LLM is the speech recognition output from whisper-large-v2, which includes both an encoder (used as the speech encoder in BLSP-KD) and a decoder (not used in BLSP-KD). It is important to note that the speech training data for BLSP-KD is significantly smaller than that for Whisper models when comparing BLSP-KD models to this baseline.


\paragraph{CFormer+LLM} This is a pipelined baseline sharing the identical model architecture as BLSP-KD. The input to the LLM is the speech recognition output from a recognition model that includes a frozen speech encoder and a CFormer adapter, both identical to those in BLSP-KD. The CFormer adapter is optimized to predict the ground-truth transcript $\mathbf{x}$ from speech input $\mathbf{s}$ using an ASR loss $\ell_\text{ASR}(\mathbf{s}, \mathbf{x})$, based on the same ASR training data described in Section~\ref{sec:training_data}.


\paragraph{BLSP} This is the end-to-end BLSP approach described in~\cite{wang2023blsp}, which uses a convolution-based subsampler as the modality adapter (denoted CNN in experiments)  and is optimized using the cross-entropy loss $\ell^\text{resp}_\text{CE}(\mathbf{s}, \mathbf{y})$ on the continuation writing instruction data synthesized from ASR training data. To ensure a fair comparison with BLSP-KD models, the BLSP baseline in our experiments also uses Qwen-7B-chat as the LLM and the encoder part of Whisper-large-v2 as the speech encoder, and is trained on the same continuation writing data as described in Section~\ref{sec:training_data}.

\section{Results and Analysis}
\begin{table*}[htbp]
\centering
\scriptsize
\renewcommand{\arraystretch}{1.3}
\begin{tabular}{lcccccccccc}
    \toprule
    \multirow{2}{*}{\textbf{System}} & \multirow{2}{*}{\textbf{Adapter}} & \multicolumn{3}{c}{\textbf{Tunable}} & \multirow{2}{*}{\textbf{Training Loss}} &  \multirow{2}{*}{\textbf{Training Data}} & \multicolumn{2}{c}{\textbf{CoVoST}} & \multicolumn{2}{c}{\textbf{MUST-C}}\\ 
    \cmidrule(lr){3-5} \cmidrule(lr){8-9} \cmidrule(lr){10-11}
     & & Whisper & Adapter & LLM & & & BLEU & Self & BLEU & Self\\
     \toprule
    Text+LLM &  N/A &   &  &  &  & & 27.7 & 100.0 & 26.4 & 100.0 \\
    Whisper+LLM  & N/A & & & & & & 24.3 & \phantom{0}58.6 & 23.3 & \phantom{0}70.2  \\
    CFormer+LLM & CFormer &  & \checkmark & & $\ell_\text{CIF},\ell_\text{ASR}$ & ASR & 20.9 & \phantom{0}45.9 & 19.6 & \phantom{0}51.6 \\
    \midrule
    BLSP & CNN &  & \checkmark & & $\ell_\text{CE}^\text{resp}$ & CW & 20.1 & \phantom{0}39.2 & 19.1 & \phantom{0}45.7 \\
    BLSP-KD\ding{192} & CNN &  & \checkmark & & $\ell_\text{KL}^\text{resp}$ & CW & 21.5 & \phantom{0}44.1 & 19.7 & \phantom{0}49.2 \\
    BLSP-KD\ding{193} & CFormer &  & \checkmark & & $\ell_\text{CIF},\ell_\text{CE}^\text{resp}$ & CW & 20.6 & \phantom{0}41.2 & 19.7 & \phantom{0}49.7 \\
    BLSP-KD\ding{194} & CFormer &  & \checkmark & & $\ell_\text{CIF},\ell_\text{KL}^\text{resp}$ & CW & 22.0 & \phantom{0}46.2 & 19.9 & \phantom{0}52.6 \\
    BLSP-KD\ding{195} & CFormer &  & \checkmark & & $\ell_\text{CIF},\ell_\text{KL}^\text{input}$ & ASR & 21.2 & \phantom{0}45.0 & 20.1 & \phantom{0}52.2 \\
    BLSP-KD\ding{196} & CFormer &  & \checkmark & & $\ell_\text{CIF},\ell_\text{KL}^\text{input},\ell_\text{KL}^\text{resp}$ & ASR,CW & 22.0 & \phantom{0}47.2 & 20.3 & \phantom{0}53.6 \\
    BLSP-KD\ding{197} & CFormer & \checkmark & \checkmark & & $\ell_\text{CIF},\ell_\text{KL}^\text{input},\ell_\text{KL}^\text{resp}$ & ASR,CW & 22.5 & \phantom{0}48.9 & 20.0 & \phantom{0}53.1 \\
    BLSP-KD\ding{198} & CFormer & \checkmark & \checkmark & \checkmark & $\ell_\text{CIF},\ell_\text{KL}^\text{input},\ell_\text{KL}^\text{resp}$ & ASR,CW & 22.7 & \phantom{0}49.4 & 20.5 & \phantom{0}54.1 \\
    \bottomrule
\end{tabular}
\caption{BLSP-KD results on zero-shot speech-to-text translation. The BLEU scores for the ST test sets are averaged across multiple translation directions. The ``Self'' column indicates the use of output generated by Text+LLM as reference.}
\label{tab:ST_results}
\end{table*}


The main evaluations are conducted on speech translation and general question-answering tasks. Given that our models do not rely on any task-specific training data, all evaluations are conducted in a \emph{zero-shot} setting. Our results indicate that the proposed BLSP-KD approach significantly surpasses the previous BLSP method in all evaluated tasks and also achieves markedly better performance than the comparable cascaded baseline, CFormer+LLM, which shares the same model architecture and training data. We will analyze the capability of BLSP models in speech recognition in Section~\ref{sec:eval_asr}.

\subsection{Speech Translation}
\label{sec:eval_st}
Table~\ref{tab:ST_results} presents the main speech translation results for the in-domain CoVoST 2.0 test sets \citep{wang2020covost}, averaged across seven English-to-X directions, and for the out-of-domain MUST-C test sets \citep{di2019must}, averaged across eight English-to-X directions\footnote{{Detailed results for each language pair are provided in the Appendix~\ref{app:details}}.}. Seven configurations of BLSP-KD are shown, each utilizing different adapter types, tunable modules, and training losses. Here are three key observations.

First, when only the adapter is tuned, all variations of the BLSP-KD approach, numbered \ding{192} to \ding{196} based on different combinations of adapter type and training loss, outperform the BLSP baseline. The highest-performing BLSP-KD\ding{196} model achieves an average BLEU improvement of 1.9 on CoVoST and 1.2 on MUST-C. This superior performance can be attributed to two primary factors:

\begin{itemize}
\item The KL loss is significantly better than the CE loss for modality alignment. This is evident from the comparison between BLSP-KD\ding{192} and BLSP (KL vs. CE with CNN as the modality adapter), and also between BLSP-KD\ding{194} and BLSP-KD\ding{193} (KL vs. CE with CFormer as the modality adapter).
\item CFormer is better than CNN for modality alignment as it addresses the length discrepancy between speech and text inputs. This advantage is demonstrated in the comparison between BLSP-KD\ding{193} and BLSP (CFormer vs. CNN with CE loss), as well as between BLSP-KD\ding{194} and BLSP-KD\ding{192} (CFormer vs. CNN with KL loss).
\end{itemize}
Additionally, an important benefit of employing CFormer is its ability to train directly on ASR data using the $\ell^\text{input}_\text{KL}$ loss without the need for any speech instruction data. This is evidenced by the competitive performance of BLSP-KD\ding{195}. The best performing model, BLSP-KD\ding{196}, utilizes the CFormer adapter and is trained on both the ASR data and the CW data.

Second, additional improvement can be achieved by fine-tuning the speech encoder and the LLM (with PLoRA adaptation). Specifically, on top of the fine-tuned BLSP-KD\ding{196} model, we unfreeze the speech encoder and fine-tune the parameters along the adapter. The resulting BLSP-KD\ding{197} model achieves an additional +0.5 BLEU improvement on the in-domain CoVoST test set. However, this results in -0.3 BLEU degradation on the out-of-domain MUST-C test sets, possibly due to overfitting the speech encoder to in-domain speech data. We then further unfreeze the parameters of the LLM through PLoRA adaptation. This resulted in the BLSP-KD\ding{198} model, which achieved an additional +0.2 and +0.5 BLEU improvement over the BLSP-KD\ding{197} model. Compared to fine-tuning the adapter only as in BLSP-KD\ding{196}, BLSP-KD\ding{198} achieves a +0.7 and +0.2 BLEU improvement overall on CoVoST and MUST-C, respectively.


Third, despite the improvements achieved by the BLSP-KD approach, its performance, as achieved by the BLSP-KD\ding{198} model, still falls short of the Whisper+LLM approach by 1.6 BLEU on CoVoST and 2.8 BLEU on MUST-C. However, this comparison between cascaded and end-to-end approaches is not entirely fair, as the Whisper ASR model in the cascaded approach includes a decoder component that is trained together with the encoder on a significantly larger amount of ASR training data (680k hours) compared to the ASR training data (3k hours) used for training BLSP-KD. The cascaded baseline CFormer+LLM is a more comparable baseline to BLSP-KD as it shares the same model architecture and utilizes the same ASR data for training. Compared to CFormer+LLM, the BLSP-KD approach achieves +1.1/+0.7 BLEU improvement on CoVoST/MUST-C with BLSP-KD\ding{196}, and +1.8/+0.9 BLEU improvement with BLSP-KD\ding{198}. This suggests that the end-to-end modeling approach employed in BLSP-KD can achieve better performance than a comparable cascaded baseline based on the same amount of ASR training data. It would be interesting to investigate whether BLSP-KD could benefit from more training data and surpass the strong cascaded Whisper+LLM baseline. We leave this scaling work for future investigations.


\subsection{General QA}
\label{sec:eval_qa}

\begin{table*}[htbp]
\centering
\scriptsize
\renewcommand{\arraystretch}{1.3}
\begin{tabular}{lcccccccc}
    \toprule
    \multirow{2}{*}{\textbf{System}} & \multirow{2}{*}{\textbf{Adapter}} & \multicolumn{3}{c}{\textbf{Tunable}} & \multirow{2}{*}{\textbf{Training Loss}} &  \multirow{2}{*}{\textbf{Training Data}} & \multirow{2}{*}{\textbf{Self-BLEU}} & \multirow{2}{*}{\textbf{Self-RougeL}}\\ 
    \cmidrule(lr){3-5}
     & & Whisper & Adapter & LLM & & & & \\
     \toprule
    Text+LLM &  N/A &   &  &  &  & & 100.0 & 100.0 \\
    Whisper+LLM  & N/A & & & & & & \phantom{0}50.3 & \phantom{0}69.5 \\
    CFormer+LLM & CFormer &  & \checkmark & & $\ell_\text{CIF},\ell_\text{ASR}$ & ASR & \phantom{0}45.2 & \phantom{0}65.2 \\
    \midrule
    BLSP & CNN &  & \checkmark & & $\ell_\text{CE}^\text{resp}$ & CW & \phantom{0}42.4 & \phantom{0}64.0 \\
    BLSP-KD\ding{192} & CNN &  & \checkmark & & $\ell_\text{KL}^\text{resp}$ & CW & \phantom{0}46.1 & \phantom{0}66.8 \\
    BLSP-KD\ding{193} & CFormer &  & \checkmark & & $\ell_\text{CIF},\ell_\text{CE}^\text{resp}$ & CW & \phantom{0}40.4 & \phantom{0}63.3 \\
    BLSP-KD\ding{194} & CFormer &  & \checkmark & & $\ell_\text{CIF},\ell_\text{KL}^\text{resp}$ & CW & \phantom{0}45.8 & \phantom{0}66.5 \\
    BLSP-KD\ding{195} & CFormer &  & \checkmark & & $\ell_\text{CIF},\ell_\text{KL}^\text{input}$ & ASR & \phantom{0}44.4 & \phantom{0}65.1 \\
    BLSP-KD\ding{196} & CFormer &  & \checkmark & & $\ell_\text{CIF},\ell_\text{KL}^\text{input},\ell_\text{KL}^\text{resp}$ & ASR,CW & \phantom{0}45.8 & \phantom{0}66.3 \\
    BLSP-KD\ding{197} & CFormer & \checkmark & \checkmark & & $\ell_\text{CIF},\ell_\text{KL}^\text{input},\ell_\text{KL}^\text{resp}$ & ASR,CW & \phantom{0}47.1 & \phantom{0}67.1 \\
    BLSP-KD\ding{198} & CFormer & \checkmark & \checkmark & \checkmark & $\ell_\text{CIF},\ell_\text{KL}^\text{input},\ell_\text{KL}^\text{resp}$ & ASR,CW & \phantom{0}46.3 & \phantom{0}66.8 \\
    \bottomrule
\end{tabular}
\caption{BLSP-KD results on zero-shot general QA tasks. Self-BLEU and self-RougeL are calculated by considering the output generated by Text+LLM as the reference.}
\label{tab:QA_results}
\end{table*}

We also evaluate the performance of the BLSP-KD approach on general question-answering (QA) tasks, following the methodology adopted in \citep{wang2023blsp}. To address the lack of a diverse instruction evaluation dataset for natural speech, the authors proposed constructing a speech instruction evaluation dataset from ASR data. This was achieved by selecting 1,460 samples from the GigaSpeech test set and employing ChatGPT to construct an appropriate question for each sample based on its transcript. Subsequently, ChatGPT was utilized to determine if a response generated by the evaluated models was acceptable, computing the acceptance rate as the evaluation metric.



However, through manual inspection, we observe that ChatGPT cannot produce reliable assessments of response quality among the various model configurations in our study. To address this, we designed alternative metrics by treating the LLM's response to a given text as the reference and calculating the SacreBLEU \citep{post-2018-call} and RougeL \citep{lin-2004-rouge} scores between different model outputs and this reference. These metrics, termed Self-BLEU and Self-RougeL, assess the difference in model behavior when using speech as input compared to using the reference transcript as input. A similar Self-BLEU metric has been used for evaluating speech translation quality, showing a similar trend among the models compared to the original BLEU metric, as shown in Table~\ref{tab:ST_results}. The results for the QA task are presented in Table~\ref{tab:QA_results}.

We summarize two main observations consistent with those from the speech translation task. First, the KL loss significantly improves QA performance, regardless of the modality adapter used. Second, most BLSP-KD models achieve better performance than the comparable cascaded baseline CFormer+LLM, with the best-performing BLSP-KD\ding{197} model outperforming it by +1.9 in both Self-BLEU and Self-RougeL. However, we note two deviations: the use of the CFormer adapter results in worse performance than the CNN adapter (see BLSP-KD\ding{193} vs. BLSP, BLSP-KD\ding{194} vs. BLSP-KD\ding{192}); and unfreezing the speech encoder (BLSP-KD\ding{197}) remains beneficial, but unfreezing the LLM (through PLoRA, BLSP-KD\ding{198}) does not.

\subsection{Speech Recognition}
\label{sec:eval_asr}

Finally, we examine the BLSP-KD approach's ability to recognize lexical details in speech as evaluated in a typical speech recognition task, following the setup in \citep{wang2023blsp}. While BLSP-KD models rely on ASR training data in the form of speech and transcription pairs, they are trained using the next token prediction task via knowledge distillation, instead of learning to directly recognize words. Despite this, these models can be instructed to perform a speech recognition task by a prompt: ``\emph{Please repeat the following words.}'' For a cascaded system, the speech recognition task can be performed either directly using the standalone ASR component or by prompting the LLM to repeat the recognized words. 
As shown in Table~\ref{tab:ASR_results}, BLSP-KD models are able to perform the speech recognition task via prompting and achieve WER scores in the range of 9.5\% to 12.1\%, which are decent but lag behind the standalone ASR systems, e.g., 5.4\% for the ASR component of the CFormer+LLM baseline and 2.8\% for Whisper. While part of the difference is due to the LLM's imperfect ability to follow the repeat prompt, as evidenced by the higher WER rate on cascaded baselines via prompting, a significant gap remains.

\begin{table}[tp]
\centering
\footnotesize
\renewcommand{\arraystretch}{1.3}
\begin{tabular}{lccc}
    \toprule
    \textbf{Method} & \textbf{Training Loss} &  \textbf{ASR} & \textbf{Prompting} \\
    \toprule
    Text+LLM    &  & 0.0 & 1.5  \\
    Whisper+LLM &  & 2.8 & 3.2   \\
    CFormer+LLM & $\ell_\text{ASR}$ & 5.4 & 6.5  \\
    \midrule
    BLSP        & $\ell_\text{CE}^\text{resp}$ & - & 10.1  \\ 
    BLSP-KD\ding{194} &  $\ell_\text{CIF},\ell_\text{KL}^\text{resp}$ & - & 10.3 \\
    BLSP-KD\ding{195} &  $\ell_\text{CIF},\ell_\text{KL}^\text{input}$ & - & 12.1 \\
    BLSP-KD\ding{196} &  $\ell_\text{CIF},\ell_\text{KL}^\text{input},\ell_\text{KL}^\text{resp}$ & - & 9.5 \\
    
    \bottomrule
\end{tabular}
\caption{Main speech recognition results measured in WER on Librispeech clean test set.}
\label{tab:ASR_results}
\end{table}

Does the gap in ASR performance suggest that BLSP-KD models cannot capture as much lexical detail as standalone ASR models? To answer this question, we designed an ASR probing experiment. Specifically, we freeze the speech encoder, adapter, and LLM, and train a 2-layer transformer as a probe to reconstruct the reference transcript from the hidden states in the BLSP-KD models. As shown in Table~\ref{tab:ASR_prob}, this probe can be inserted at any layer of the LLM\footnote{For example, the 0-th layer represents the hidden states at the output of the modality adapter or the input layer of the LLM, and the 32nd layer represents the hidden states at the final layer of the LLM.}. We observe that the recognition results obtained by the probe are much lower than those obtained by prompting the LLM, with the lowest WER score of 4.2\% obtained from BLSP-KD\ding{196}, which is lower than the WER score of 5.4\% of the ASR component of the CFormer+LLM baseline. Additionally, we observe that the recognition results obtained at different layers are similar and that the $\ell_\text{KL}^\text{input}$ loss (BLSP-KD\ding{195}) is helpful for capturing lexical details, resulting in much lower WER scores than using the $\ell_\text{KL}^\text{resp}$ loss (BLSP-KD\ding{194}). This result suggests that our BLSP-KD models are indeed able to capture significant lexical details despite not being trained on the ASR task.


\begin{table}[tp]
\centering
\footnotesize
\renewcommand{\arraystretch}{1.3}
\begin{tabular}{lccccc}
    \toprule
    \multirow{2}{*}{\textbf{Method}} & \textbf{0-th} &  \textbf{8-th} & \textbf{16-th} & \textbf{24-th} & \textbf{32-th} \\
     & \textbf{layer} & \textbf{layer} & \textbf{layer} & \textbf{layer} & \textbf{layer} \\
    \toprule
    BLSP-KD\ding{194} & 6.9 & - & - & - & 6.8   \\
    BLSP-KD\ding{195}   & 4.6 & 4.4 & 4.3 & 4.3 & 4.4  \\
    BLSP-KD\ding{196} & 4.3 & - & - & - & 4.2 \\
    
    \bottomrule
\end{tabular}
\caption{The results of inserting the ASR probe into different layers of the LLM, measured in WER on Librispeech clean test set.}
\label{tab:ASR_prob}
\end{table}

\section{Related Works}

\paragraph{Decoder-Only Speech-Text Modeling} Large Language Models (LLMs) have achieved remarkable performance on various natural language processing tasks \citep{openai2023gpt,touvron2023llama}. Some research aims to incorporate speech signals into decoder-only pre-trained transformers for accomplishing spoken language tasks such as speech recognition \citep{fathullah2023prompting}, speech translation \citep{wu2023decoder}, and speech understanding \citep{wang2023speech}. In addition to task-specific models, there are unified models that combine text and speech modalities for all speech processing tasks. Models like AudioPaLM \citep{rubenstein2023audiopalm} and SpeechGen \citep{wu2023speechgen} model discrete audio tokens and text tokens as a shared vocabulary, but they may suffer from the information loss due to the quantization. To preserve more acoustic information, VIOLA \citep{wang2023viola} employs codec code instead of hidden units from HuBERT to discretize speech, while LauraGPT \citep{chen2023lauragpt} combines continuous and discrete features for audio signals. However, these models trained via multi-task learning predominantly focus on downstream speech processing tasks, neglecting the potential of LLMs to unlock new task and zero-shot capabilities.

\paragraph{Interact with LLMs through Speech} Several studies have focused on combining specialized speech models with LLMs to enable speech-based interactions. Initial efforts in this field, such as HuggingGPT \citep{shen2023hugginggpt} and AudioGPT \citep{huang2023audiogpt}, employed a cascading model structure, linking LLMs with additional ASR models for speech input. However, these models exhibited high complexity, error accumulation, and loss of paralinguistic loss. Recent works have started to explore end-to-end model architectures. SpeechGPT \citep{zhang2023speechgpt} takes the discretized output of HuBERT as a specialized linguistic unit, but faced challenges in achieving multiple rounds of dialogue due to the high sampling frequency of the discrete unit. LLaSM \citep{shu2023llasm} and SLM \citep{wang2023slm} has constructed an extensive speech instruction dataset intended for training the modality adapter. Their approach heavily rely on the TTS synthesized instruction data, resulting in limited robustness across different speakers. COSMIC \citep{pan2023cosmic} and WavLLM \citep{hu2024wavllm} addresses this by utilizing ChatGPT to query and generate responses for the transcripts in ASR data. BLSP \citep{wang2023blsp} eliminates the dependence on speech instruction data by transferring instruction following capabilities from LLMs to the speech modality via behavior alignment of corresponding speech and transcript.


\section{Conclusion}
Our proposed BLSP-KD method, Bootstrap Language-Speech Pretraining via Knowledge Distillation, addresses the challenges of extending large language models (LLMs) to speech inputs. It optimizes speech-text alignment by minimizing divergence in next-token predictions and segments speech into one-to-one text tokens using a continuous-integrate-and-fire strategy. Additionally, Partial LoRA (PLoRA) supports LLM finetuning for speech inputs under knowledge distillation. Quantitative evaluation shows that BLSP-KD significantly outperforms cascaded systems and previous end-to-end baselines, enhancing LLMs' ability to follow instructions from speech inputs. We hope that, by aligning large language models with the speech modality, this approach can be further extended to capture paralinguistic cues in speech inputs and generate expressive speech as responses, opening new possibilities for spoken language interactions.

\bibliography{custom}

\appendix

\section{Detailed Results for Speech-to-Text Translation}
\label{app:details}

We present the performance of speech translation in each direction, as illustrated in Tables \ref{tab:covost} and Table \ref{tab:mustc}. 
For CoVoST-2, we evaluate our method on seven translation directions: English (en) to Catalan (ca), German (de), Indonesian (id), Japanese (ja), Slovenian (sl), Swedish (sv) and Chinese (zh).
Additionally, we conduct experiments on MUST-C for all eight translation directions: English (en) to German (de), Spanish (es), French (fr), Italian (it), Dutch (nl), Portuguese (pt), Romanian (ro), and Russian (ru).

\begin{table*}[tbp]
\centering
\footnotesize
\begin{tabular}{l|ccccccc}
    \toprule
    \textbf{Method} & \textbf{en-ca} & \textbf{en-de} & \textbf{en-id} & \textbf{en-ja}& \textbf{en-sl} & \textbf{en-sv} & \textbf{en-zh} \\
    \midrule
    Text+LLM & 30.0 & 28.9 & 19.4 & 24.2 & 13.8 & 30.2 & 47.6 \\
    Whisper+LLM & 25.3 & 25.3 & 15.4 & 22.7	 & 11.9 & 25.8 & 43.6  \\
    CFormer+LLM & 21.7 & 22.1 & 12.2 & 20.5 & \phantom{0}9.9 & 22.3 & 37.8 \\
    \midrule
    BLSP & 21.1 & 21.1 & 10.9 & 19.5 & 9.1 & 21.4 & 38.0 \\
    BLSP\ding{192} & 22.2 & 23.0 & 12.3	& 20.2 & 10.2 & 23.0 & 39.3\\
    BLSP\ding{193} & 21.8 & 22.1 & 11.2 & 19.1 & \phantom{0}9.5 & 22.2 & 38.4 \\
    BLSP\ding{194} & 22.8 & 23.7 & 12.7 & 21.0 & 10.3& 23.4 & 39.9\\
    BLSP\ding{195} & 21.9 & 22.7 & 12.2 & 20.5 & \phantom{0}9.7 & 22.5 & 39.1\\
    BLSP\ding{196} & 22.7 & 23.4 & 12.8 & 21.0 & 10.4 & 23.5 & 40.3\\
    BLSP\ding{197} & 23.3 & 24.2 & 13.3 & 21.3 & 10.5 & 24.0 & 40.9 \\
    BLSP\ding{198} & 23.5 & 24.4 & 13.4 & 21.3 & 10.7 & 24.5 & 41.3 \\
    \bottomrule
\end{tabular}
\caption{ST results on CoVoST-2.}
\label{tab:covost}
\end{table*}

\begin{table*}[tbp]
\centering
\footnotesize
\begin{tabular}{l|cccccccc}
    \toprule
    \textbf{Method} & \textbf{en-de} & \textbf{en-es} & \textbf{en-fr} & \textbf{en-it} & \textbf{en-nl} & \textbf{en-pt} & \textbf{en-ro} & \textbf{en-ru} \\
    \midrule
    Text+LLM & 26.2 & 31.9 & 39.4 & 23.6 & 26.9 & 27.3 & 20.5 & 15.3 \\
    Whisper+LLM & 22.8 & 27.9 & 35.3 & 20.9 & 22.9 & 24.1 & 18.4	 & 13.9\\
    CFormer+LLM & 19.2 & 24.1 & 29.0 & 17.7	& 19.1 & 20.4 & 15.6	 & 11.9 \\
    \midrule
    BLSP & 19.2 & 23.5 & 29.1 & 16.8 & 18.9	& 18.8 & 15.1 & 11.3 \\
    BLSP\ding{192} & 19.7& 24.2 & 28.9 & 17.1 & 19.5& 21.2 & 14.8 & 12.0 \\
    BLSP\ding{193} & 19.3 & 24.0 & 29.1 & 17.5 & 19.9& 21.3 & 15.3	& 11.6 \\
    BLSP\ding{194} & 19.9 & 24.5 & 30.0 & 17.3 & 19.5 & 21.0 & 15.3 & 11.9 \\
    BLSP\ding{195} & 20.2 & 24.2 & 30.4 & 17.7 & 19.7& 21.3 & 15.5	& 12.1 \\
    BLSP\ding{196} & 20.9 & 24.7 & 31.1 & 17.9 & 19.7 & 21.4 & 15.2 & 11.9 \\
    BLSP\ding{197} & 20.3 & 24.6 & 29.8	 & 17.7 & 19.7 & 21.1 & 15.4 & 11.7 \\
    BLSP\ding{198} & 20.7 & 24.8 & 31.4	 & 18.1	 & 20.2	& 21.7 & 15.3 & 11.7 \\
    \bottomrule
\end{tabular}
\caption{ST results on MUST-C.}
\label{tab:mustc}
\end{table*}

\end{document}